%% file: acl2023.tex
\pdfoutput=1

\documentclass[11pt]{article}

\usepackage[final]{acl2023}

\usepackage{times}
\usepackage{latexsym}

\usepackage[T1]{fontenc}

\usepackage[utf8]{inputenc}

\usepackage{microtype}

\usepackage{inconsolata}

\usepackage{graphicx}
\usepackage[scaled=0.85]{beramono}

\input{commands}

\title{Do These LLM Benchmarks Agree? \\ 
Fixing Benchmark Evaluation with \benchbench}

\author{\normalsize
Yotam Perlitz$^1$ \qquad Ariel Gera$^1$ \qquad Ofir Arviv$^1$ \qquad Elron Bandel$^1$ \\ 
\textbf{ \normalsize Asaf Yehudai}$^1$ \quad \textbf{ \normalsize Eyal Shnarch}$^1$ \quad \textbf{ \normalsize Michal Shmueli-Scheuer}$^1$ \quad \textbf{ \normalsize Leshem Choshen}$^{2,3}$ 
\\ \\
$^1$IBM Research AI \quad $^2$MIT CSAIL \quad $^3$MIT-IBM\\
\texttt{\{y.perlitz,leshem.choshen\}@ibm.com}}

\begin{document}
\maketitle

\input{content}

\bibliography{custom,anthology}

\input{appendix}

\end{document}

%% file: commands.tex
\usepackage{multirow}

\newcommand{\cready}[1]{}

\newcommand\benchvalidname{\texttt{BAT}}
\newcommand\benchbench{\texttt{BenchBench}}
\newcommand\benchbenchbold{\texttt{\textbf{BenchBench}}}

\usepackage{etoolbox}
\usepackage{pgf} 
\definecolor{lowcolor}{HTML}{44ff44} 
\definecolor{midcolor}{HTML}{ffffff}  
\definecolor{highcolor}{HTML}{ff0000}  
\newcommand*{\opacity}{60}
\newcommand*{\minval}{0.10}
\newcommand*{\midval}{0.19} 
\newcommand*{\maxval}{0.31}
\newcommand{\ccell}[1]{
    \ifdimcomp{#1pt}{>}{\maxval pt}{#1}{
        \ifdimcomp{#1pt}{<}{\minval pt}{#1}{
          \ifdimcomp{#1pt}{<}{\midval pt}{
            \pgfmathparse{int(round(100*(#1/(\midval-\minval))-(\minval*(100/(\midval-\minval)))))}\xdef\tempa{\pgfmathresult}\cellcolor{midcolor!\tempa!lowcolor!\opacity}#1}{
            \pgfmathparse{int(round(100*(#1/(\maxval-\midval))-(\midval*(100/(\maxval-\midval)))))}\xdef\tempa{\pgfmathresult}\cellcolor{highcolor!\tempa!midcolor!\opacity}#1}}}}


%% file: content.tex
\begin{abstract}
Recent advancements in Language Models (LMs) have catalyzed the creation of multiple benchmarks.
A crucial task, however, is assessing the validity of the benchmarks themselves. 
This is most commonly done via Benchmark Agreement Testing (\benchvalidname{}), where new benchmarks are validated against established ones using some agreement metric (e.g., Spearman correlation). 
Despite the crucial role of \benchvalidname{} for benchmark builders and consumers, 
there are no standardized procedures for such agreement testing. 
This deficiency can lead to invalid conclusions, fostering mistrust in benchmarks and upending the ability to choose the appropriate benchmark. 
By analyzing over 50 prominent benchmarks, we demonstrate how some overlooked methodological choices can significantly influence \benchvalidname{} results, potentially undermining the validity of conclusions. 
To address these inconsistencies, we propose a set of best practices for \benchvalidname{} and demonstrate how utilizing these methodologies greatly improves \benchvalidname{} robustness and validity.
To foster adoption and facilitate future research, we introduce \benchbench{}\footnote{\url{https://github.com/IBM/benchbench}}, a Python package for \benchvalidname{}, and release the \benchbench{}-leaderboard\footnote{\url{https://hf.co/spaces/ibm/benchbench}}, a meta-benchmark designed to evaluate benchmarks using their peers. 

\end{abstract}

\begin{figure}[ht]
    \centering
    \includegraphics[width=\linewidth]{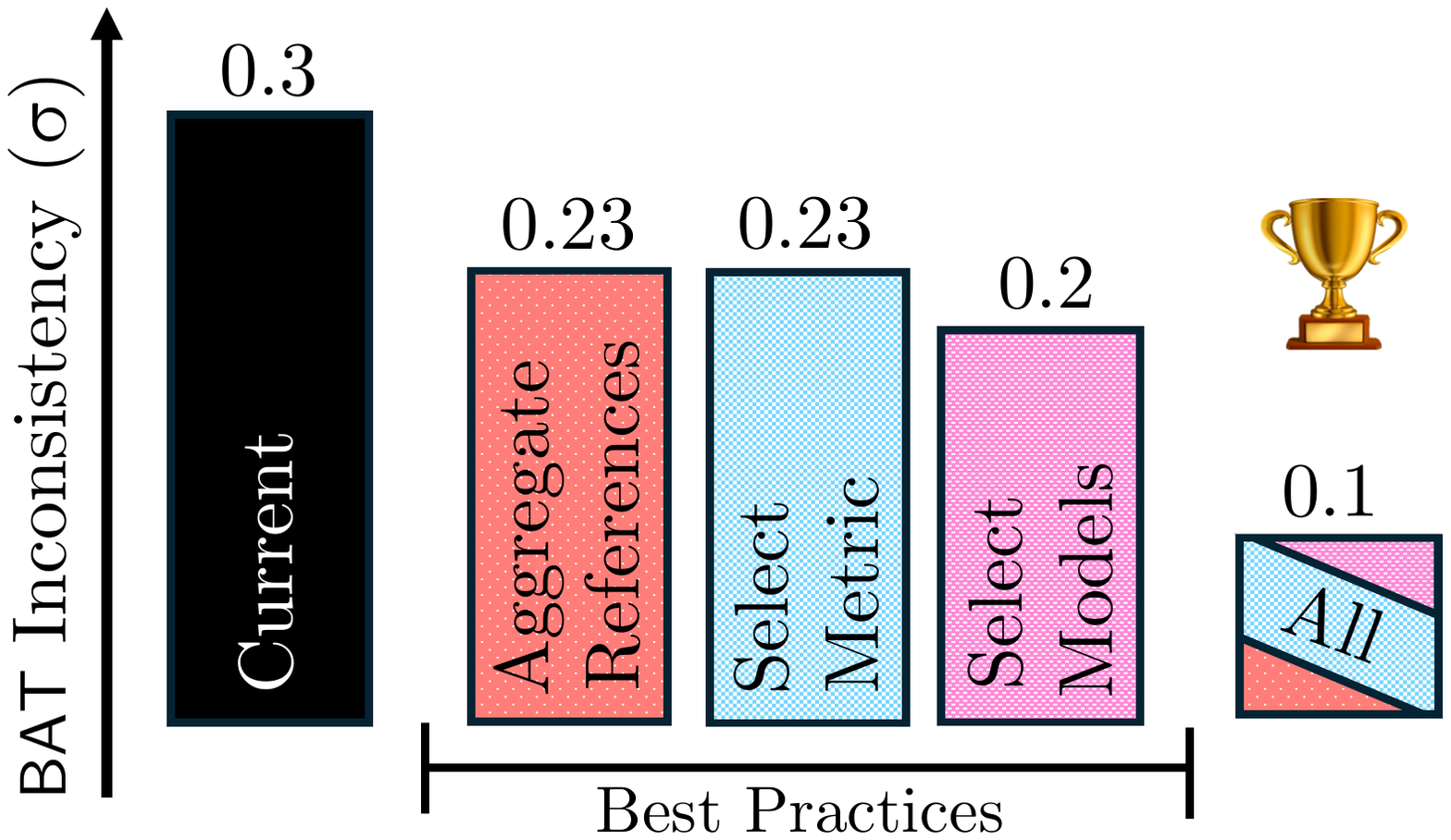}
    \caption{\textbf{Running \benchvalidname{} using our best practices increases consistency by 3x.} The average standard deviation of \benchvalidname{} results over multiple instances is drastically decreased using our best practices, without incurring further computational costs. These best practices can be easily applied using our \benchbench{} package. Further details in Table~\ref{tab:ablations}.}
    \label{fig:ablation_bar.pdf}
\end{figure}

\section{Introduction} 

As Language Models (LMs) increasingly excel across a broad range of tasks, new benchmarks -- often measuring similar abilities -- are constantly proposed. 
This deluge of benchmarks underscores the importance of \textit{Benchmark Agreement Testing} (\benchvalidname{}). \benchvalidname{} involves validating a new benchmark by comparing it against established and trusted benchmarks, using statistical agreement metrics. This comparison is based on the performance scores of models across the different benchmarks.  

\benchvalidname{} is often used to validate that a new proposed benchmark measures what it was designed to measure. 
The expectations from this measurement depend on the benchmark's goal; 
demonstrating high agreement can serve to show that a new benchmark captures model abilities similar to those measured by established and well trusted benchmarks.~\citep{Lei2023S3EvalAS, Viswanathan2023Prompt2ModelGD, Chang2023DoLM, Li2024TreeEvalBE, Prabhu2024LifelongBE, He2024UltraEvalAL}. 
High agreement can also validate that an efficient version of a benchmark (e.g., requiring less compute or labeling) measures the same thing as the original benchmark~\citep{Perlitz2023EfficientB, Polo2024tinyBenchmarksEL, Prabhu2024LifelongBE, Vivek2023AnchorPB}.
In contrast, if a benchmark aims to test a unique trait -- one that is not properly covered by existing benchmarks -- \benchvalidname{} will be used to demonstrate the disagreement of such benchmarks with existing ones~\citep{Yuan2024PRobELMPR,waldis2024holmes}. 
The above goals are relevant both for benchmark creators and for benchmark consumers. Creators will typically use \benchvalidname{} to validate the properties of their new benchmark; benchmark consumers might use it to choose which existing benchmark they want to use.

However, despite the wide application of \benchvalidname{} in recent years, there is a glaring absence of common methodology. 
Specifically, the significance of several methodological decisions in \benchvalidname{} is currently overlooked, undermining the validity of any conclusions made.

In this work, we aim to bring order and consistency into the practice of \benchvalidname{}. 
Analyzing more than $50$ of the most common benchmarks (\S\ref{benchmarks}), spanning over $200$ models, we show the critical impact of several methodological decisions in \benchvalidname{}, effectively altering the conclusions that researchers will draw from their analyses (\S\ref{sec:analysis}).

\begin{figure}[t]
    \centering
    \includegraphics[width=\linewidth]{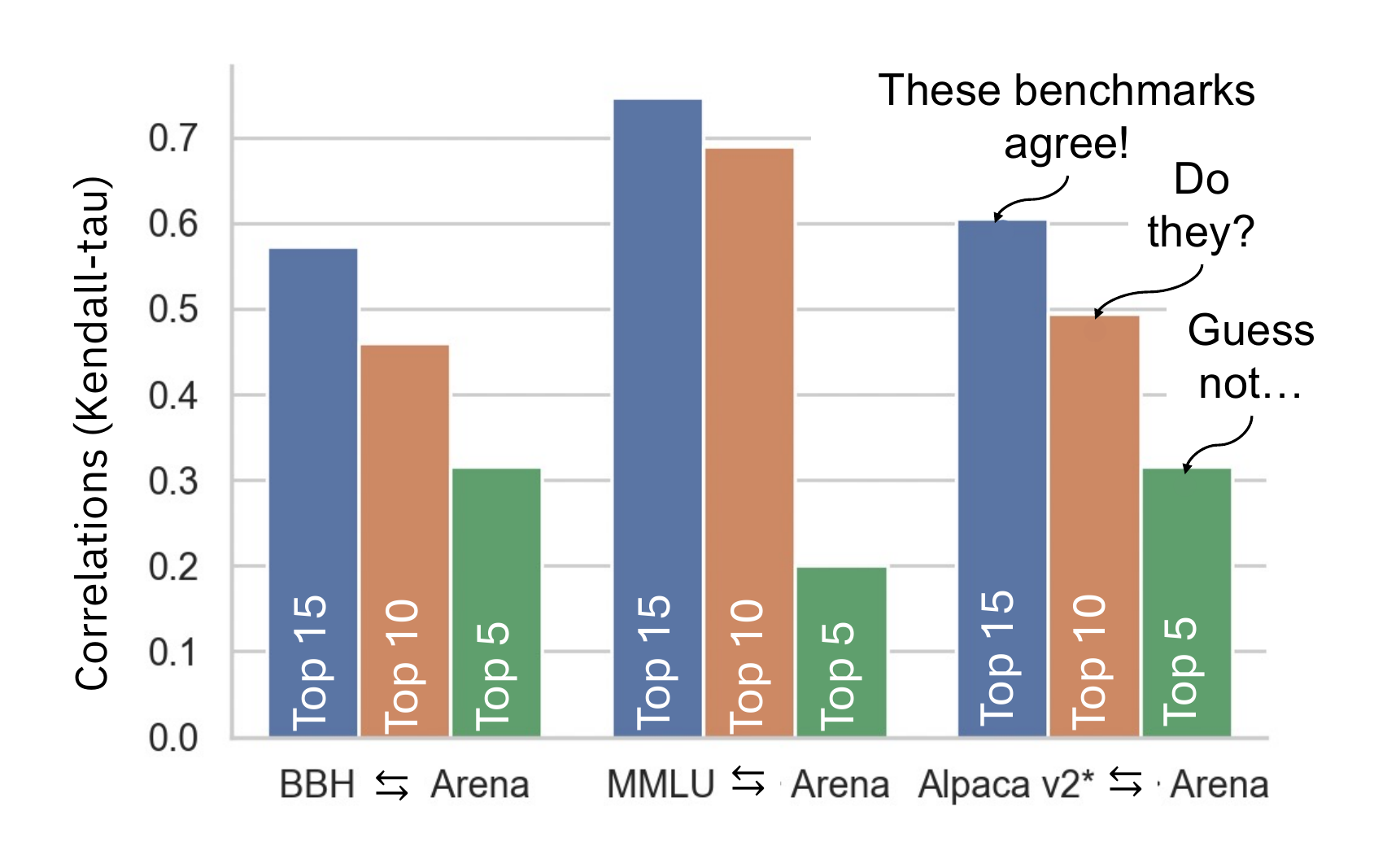}
    \caption{\textbf{\benchvalidname{} Conclusions depend on the models considered}. Kendall-tau correlations between the LMSys Arena benchmark and three other benchmarks: BBH, MMLU, and Alpaca v2. Each group of bars represents the correlation for different sets of top models, specifically the top 5, top 10, and top 15 (overlapping) models (according to the Arena). The results indicate that the degree of agreement between benchmarks varies with the number of top models considered, highlighting that different selections of models can lead to varying conclusions about benchmark agreement.}
    \label{fig:cover}
\end{figure}

We focus on three such critical choices: selecting the reference benchmark (\S\ref{subsec:reference_selection_matters}), the models included in the test (\S\ref{subsec:model_selection}), as well as the correlation metrics and their interpretation (\S\ref{subsec:corr_metric_matters}).
For example, as seen in Figure~\ref{fig:cover}, choosing a different subset of models produces substantially different correlation scores, leading to different conclusions about benchmark agreement. 
The figure demonstrates that two benchmarks can (and often do) show high agreement across a wide range of models, while agreement over a few top-ranked models remains low.

Building upon our findings, we compile a set of best practices for \benchvalidname{} (\S\ref{sec:best_practices}) and demonstrate their impact (Figure~\ref{fig:ablation_bar.pdf} and Table~\ref{tab:ablations}).  
To foster adoption and promote reproducibility, we have implemented these guidelines into \textit{\benchbench{}}, a Python package for \benchvalidname{} (\S\ref{sec:bb}). 
\benchbench{} supplies users not only with a framework but also with the data needed to perform \benchvalidname{}, relieving users of the computational and time burden of gathering multiple benchmarks for comparison. Notably, when using \benchbench{}, applying our best practices for running \benchvalidname{} will not require further computational resources.
Furthermore, \benchbench{} is built to continually evolve, allowing easy addition of new benchmarks.

Lastly (\S\ref{sec:bb}), we introduce the \benchbench{}-Leaderboard. Using \benchbench{} as its back-end, the \benchbench{}-Leaderboard is a dynamic leaderboard that provides easy access to \benchvalidname{} results for established benchmarks.
By ranking benchmarks based on their agreement with the user's desired set of reference benchmarks, the \benchbench{}-Leaderboard facilitates making informed evaluation decisions.



To sum up, our contributions are as follows:
\begin{enumerate}
    \item We perform a large-scale analysis of benchmark agreement, highlighting the impact of several crucial methodological decisions (\S\ref{sec:analysis}).
    \item We propose guidelines for reliable and standardized \benchvalidname{} (\S\ref{sec:best_practices}) and demonstrate their impact.
    \item We release \benchbench, a Python package for \benchvalidname{} implementing the guidelines and incorporating them with the required benchmark data (\S\ref{sec:bb}).
    \item We harness \benchbench{} as the back-end for a new meta-benchmark (\S\ref{sec:bb}).
\end{enumerate}

\section{Setup}
\label{benchmarks}

For our analysis, we use over $40$ benchmarks, with their results cutoff at Jan 2024. 
The benchmarks we used include: AGI Eval~\citep{Zhong2023AGIEvalAH}, Alpaca (v2)~\citep{alpaca_eval}, and its length-adjusted version~\citep{Dubois2024LengthControlledAA}, HuggingFace OpenLLM Leaderboard~\citep{open-llm-leaderboard}, MMLU~\citep{Hendrycks2020MeasuringMM}, MAGI~\citep{MAGIbenchmark}, Chatbot-Arena and MTBench~\citep{Zheng2023JudgingLW}, Big Bench Hard~\citep{suzgun2022challenging}. HumanEval~\citep{Chen2021EvaluatingLL} ARC~\citep{clark2018think}, HellaSwag~\citep{zellers2019hellaswag}, TruthfulQA~\citep{lin2022truthfulqa}, Winogrande~\citep{sakaguchi2019winogrande}, GSM8k~\citep{cobbe2021traininggsm8k}. 
EQ-Bench (v2)~\citep{paech2023eqbench}, ArenaHard~\citep{arenahard2024} and  OpenCompass~\citep{2023opencompass}.
For a wider survey of benchmarks used, see App.~\ref{app:benchmark_used}. 

Our analysis focuses on evaluating agreement between two benchmarks -- a \textit{reference benchmark} (established and commonly acceptable) and a \textit{target benchmark} (the one we assess, e.g., a new benchmark).
Specifically, agreement is calculated as the correlation over the models ranks (using Kendall~\citep{Kendall1938ANM}) or scores (using Pearson~\citep{PearsonVIINO}).

We note that an inherent constraint in \benchvalidname{} is the number of intersecting models between the benchmarks (i.e., models appearing in both benchmarks). Benchmarks lacking a sufficiently large set of intersecting models (for this work, we chose $\ge5$), cannot be reliably used for \benchvalidname{}.

\input{ablation_table}

\section{\benchvalidname{} Methodological Decisions: An Analysis} \label{sec:analysis}

When conducting \benchvalidname{}, researchers face a multitude of decisions: which reference benchmarks to compare against, which models to select for comparison, which metrics to use, how to define "agreement" between benchmarks, and so on.

In the absence of guidelines, benchmark creators often make arbitrary choices, without clear justification or consistency across different studies.

In this section, we demonstrate how such arbitrary choices hinder the validity of \benchvalidname{} conclusions. 
Next, we highlight how commonly reported \benchvalidname{} results can foster false expectations among benchmark consumers.

\begin{figure}[ht]
    \centering
    \includegraphics[width=\linewidth]{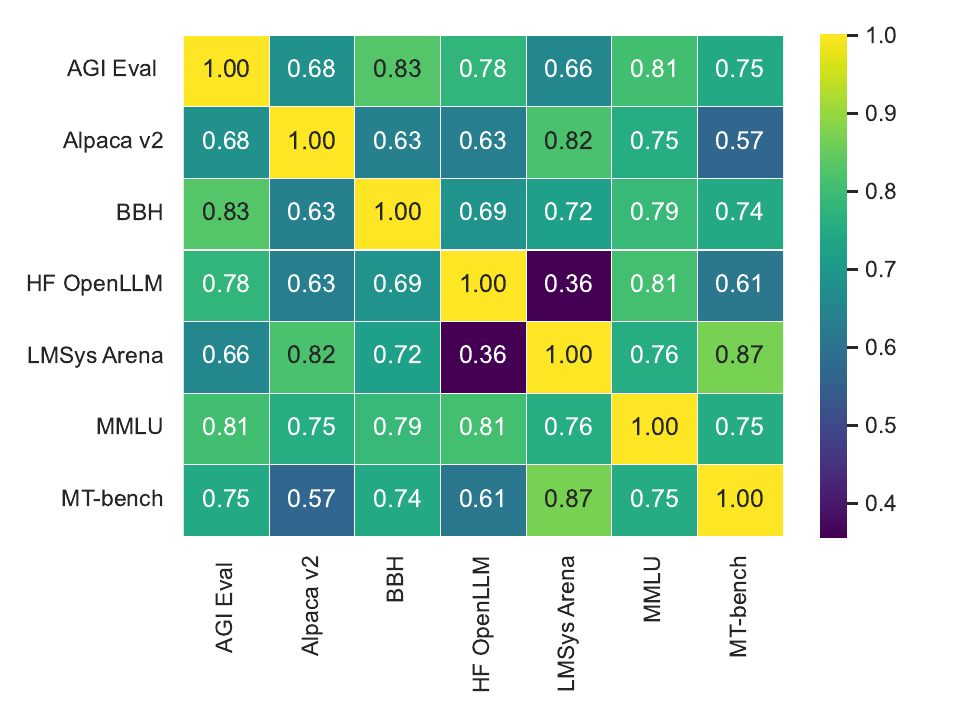}
    \caption{\textbf{Agreement scores significantly vary across different appropriate reference benchmarks}. Kendall-tau correlations between pairs of benchmarks that are seemingly valid for \benchvalidname{}. Each is taken over 20 models sampled at random.}
    \label{fig:benchmark_reference_matter}
\end{figure}

\subsection{The Choice of Reference Benchmark Matters} \label{subsec:reference_selection_matters}

Finding a reference benchmark for \benchvalidname{} is a non-trivial task. 
One needs to find a well-established benchmark, whose data is readily available, and which exhibits a large enough overlap with the models already evaluated in the target benchmark.
Due to the above difficulty, \benchvalidname{} is commonly done against one or two reference benchmarks~\citep{Yuan2024PRobELMPR}.
Benchmarks can be divided into groups according to their measured abilities -- for example, holistic benchmarks that aim to measure some loosely-defined construct of \textit{overall} model quality, such as  BigBench~\citep{srivastava2023beyond}, benchmarks measuring coding abilities ~\citep{Chen2021EvaluatingLL}, math benchmarks~\citep{cobbe2021traininggsm8k}, etc.
Thus, when selecting a reference benchmark, there is often a somewhat arbitrary choice between several possible benchmarks which are all seemingly appropriate.

Figure~\ref{fig:benchmark_reference_matter} illustrates the variability caused by such arbitrary choices: for each target benchmark, different reference benchmarks produce wildly varying agreement scores.
For example, Alpaca V2 (second row from above) demonstrates a wide range of agreement levels with other benchmarks, spanning from a mediocre agreement of $0.57$ with MT-bench to a high agreement of $0.82$ with LMSys Arena, even though both of these reference benchmarks are considered to measure similar abilities. 
This variability calls into question the validity of conclusions based on applying \benchvalidname{} when relying on a single reference benchmark.

To address this issue, we advocate using an aggregated reference benchmark that consolidates results of multiple benchmarks based on the mean-win-rate; see more on this in \S\ref{sec:aggregate}. 

\subsection{The Choice of Models Matters} \label{subsec:model_selection}

In performing \benchvalidname{}, one measures some agreement metric over the scores of a group of models overlapping between the target and reference benchmark.
Typically, authors arbitrarily pick some small set of models for their analysis. However, as we detail below, both the quantity and the properties of the selected models should be taken into account when drawing conclusions from \benchvalidname{}.

\begin{figure}[ht]
    \centering

        \includegraphics[width=\linewidth]{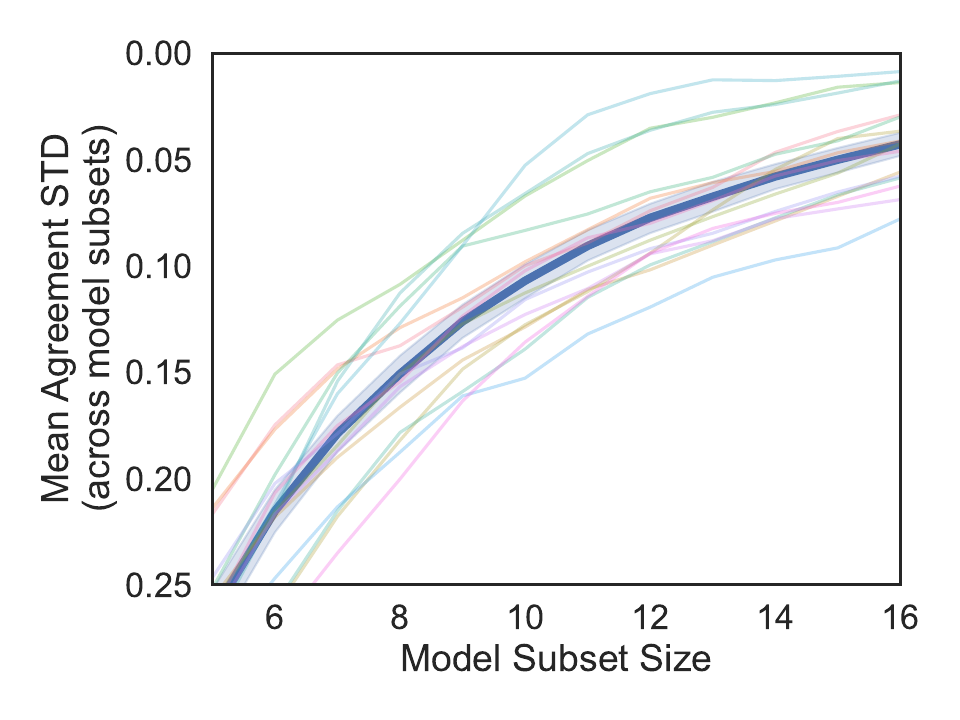} 
        \caption{\textbf{Agreement variance is inversely related to model subset size.} The mean standard deviation of the Kendall-tau correlations arising from performing \benchvalidname{} using different randomly sampled model subsets. The blue line represents the benchmark mean while the other ones are for the benchmarks listed in App~\ref{app:benchmarks_used_in_visualizations}.}
        \label{fig:n_models_matters}
\end{figure}

\paragraph{The Number of Compared Models Matters} \label{n_models_matter}
Figure~\ref{fig:n_models_matters} illustrates the relationship between the number of models and the variability of \benchvalidname{} results.
It shows that with a small amount of models, \benchvalidname{} results can get highly unreliable, with a standard deviation approaching $0.25$. 
For instance, in our analysis we found that the Kendall-tau correlation between LMSysArena and MT-Bench can range from approximately 0.65 to 0.99, depending on the particular number of models chosen. 
Thus, we see that the common practice of using a small number of models for \benchvalidname{} may jeopardise the validity of conclusions.

\paragraph{Granularity Matters}\label{subsec:Resolution matters}
Performing \benchvalidname{} produces a score that indicates high or low agreement. However, the meaning of this score will differ depending on the models included in the analysis.
For example, as seen in Figure~\ref{fig:cover}, for a given pair of benchmarks, the agreement obtained over similarly strong models will generally be lower than over a set of models of varying qualities.

\begin{figure}[t]
    \centering
    \includegraphics[width=\linewidth]{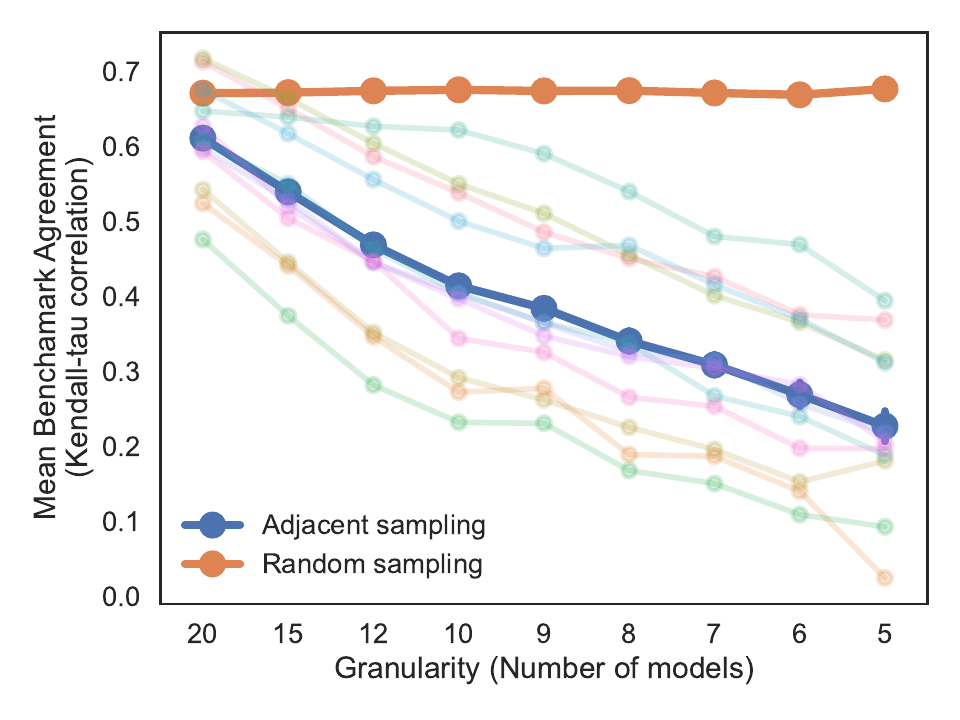}
    \caption{\textbf{Agreement is lower for closely ranked models.} Mean correlation (y) between each benchmark (lines) and the rest, given different numbers of models. The Blue and Orange lines are the average of all benchmark pair correlations with models sampled randomly (orange) or in contiguous sets (blue). The shaded lines represents adjacent sampling for the the set of benchmarks listed in App~\ref{app:benchmarks_used_in_visualizations}.}
    \label{fig:pair_agreement_vs_resolution}
\end{figure}

To quantify this phenomenon, we investigate benchmark agreement where the subset of models selected is not completely random, but is constrained to sets of models that are adjacent in rank (e.g., models 3-7)\footnote{Note that the sets of adjacent models were not selected from a specific rank location (e.g., Top, Bottom, Middle) but were randomly selected from the full range. For an analysis of such location-dependent sets, see App~\ref{subsec:Model Tier Matters}.}.
Adjacent models have more similar performance. Thus, their score differences and ranking may be less stable, resulting in lower correlation scores.
In Figure~\ref{fig:pair_agreement_vs_resolution}, we show that indeed, for a given number of models, the correlation score when considering adjacent models is lower than that of randomly sampled models, with a stronger effect as the number of models in the subset decreases.


This discrepancy emphasizes the importance of reporting \benchvalidname{} scores at multiple levels of granularity.
This would enable managing the expectations of benchmark consumers, who may expect and desire a specific level of granularity (e.g., getting the very best models right, or discriminating between strong and weak models).

\subsection{The Choice of Correlation Metric\\ (and Threshold) Matters} \label{subsec:corr_metric_matters}

\benchvalidname{} is the process of measuring correlations of model scores (or ranks) between two benchmarks. 
Once a correlation score is obtained, this score is commonly interpreted based on how it compares to some threshold;  surpassing the threshold means the agreement is considered "high", while falling below it means the agreement is "low".

Currently, there are no consistent standards for the types and thresholds of correlation metrics.
For instance, \citet{Liu2021DoQA} utilized both rank and score correlations, setting a uniform threshold of $0.8$ for both, whereas \citet{sun2023validity} exclusively employed rank correlation and opted for a distinct threshold of $0.7$.

\begin{figure}[h]
        
        \includegraphics[width=\linewidth]{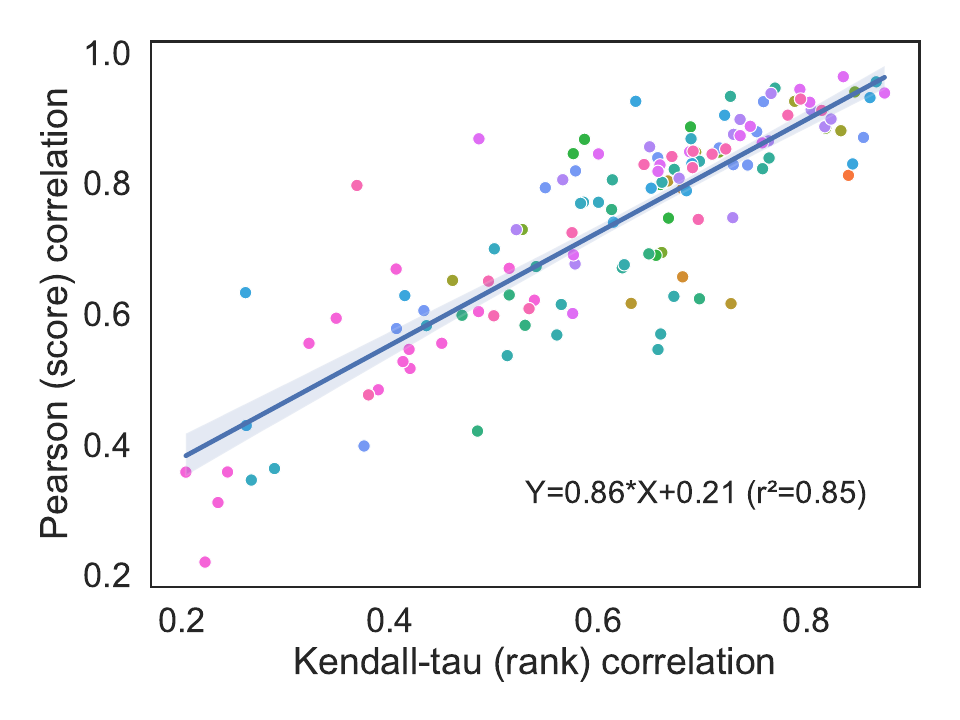} 
        \caption{\textbf{Agreement measures are linearly depended but biased.} The Kendall-tau and Pearson correlation of all benchmark pairs show a strong linear dependence, and a bias factor of $0.21$. Colors represent the different benchmarks listed in App~\ref{app:benchmarks_used_in_visualizations}.
        \label{fig:kt_pearson_correlation}}
    
\end{figure}

To improve our understanding on the significance of these choices, we analyse the relationship between rank (Kendall-tau) and score (Pearson) correlation metrics. 
In Figure~\ref{fig:kt_pearson_correlation} we present correlation scores between different pairs of benchmarks with varying model subsets. We observe a strong linear relationship ($r^{2}=0.85$) between the two correlation functions, indicating that they exhibit similar behavior in measuring agreement. However, the figure also shows a consistent score difference of approximately $0.2$ between the two metrics, indicating a potential flaw in the current practice of applying the same threshold regardless of the metric chosen. 
This underscores the necessity for a data-driven approach -- comparative in nature -- to interpret correlation scores; see \S\ref{action:corr} for more details.


\section{\benchvalidname{} Best Practices} \label{sec:best_practices}

\paragraph{Use an \textit{Aggregate} Reference Benchmark}\label{sec:aggregate}
The choice of reference benchmark can significantly affect the validity of \benchvalidname{} conclusions, as demonstrated by the variability in agreement scores when different single benchmarks are used as references (\S\ref{subsec:reference_selection_matters}, Figure~\ref{fig:benchmark_reference_matter}).
To mitigate this variability, we propose combining the results from all benchmarks appropriate for the goal of the \benchvalidname{} (e.g., benchmarks measuring similar or dissimilar abilities) into an aggregate reference benchmark by averaging their model win-rates. This approach reduces the influence of outliers and provides a more stable and robust measure of agreement, leading to more reliable conclusions.
For example, when using \benchvalidname{} to validate some efficient holistic benchmark, the reference benchmark should be the aggregate of all available holistic benchmarks.
By combining results from a group of benchmarks, the aggregate benchmark provides both a more stable and robust basis for comparison. 
Notably, since the aggregate benchmark captures the distribution of relevant results, it constitutes a better measure of the underlying construct represented by the group, called in the literature convergent validity~\citep{carlson2012understanding}.


Measuring the effect of such methodology, in Table~\ref{tab:ablations}, we compare the standard deviation of \benchvalidname{} correlation results when using arbitrary reference benchmarks (first line) to that when using the aggregate, it shows that the standard deviation of the correlation drops with our recommendation by more that $30$\%.

\paragraph{Use a Data-driven Threshold} \label{action:corr}
Using predetermined thresholds to interpret correlation scores can be misleading, as the relative nature of “high” or “low” agreement varies depending on the context, such as model granularity (\S\ref{subsec:corr_metric_matters}, Figure~\ref{fig:pair_agreement_vs_resolution}).
A more accurate and context-aware assessment can be achieved by using a data-driven approach that compares the target benchmark’s agreement with a reference benchmark (preferably an aggregate) to the distribution of agreement scores from various other benchmarks against the same reference. The steps of this approach are as follows:

\begin{enumerate}

\item \textbf{Compile a Distribution:} Begin by compiling a distribution of agreement scores from various benchmarks relative to the chosen reference benchmark. 
\item \textbf{Calculate the Target Benchmark’s Z-Score:} Next, compare the target benchmark’s correlation score to this distribution by calculating its Z-score. Indicating how the target benchmark’s agreement compares to that of other benchmarks.
\item \textbf{Interpret the Z-Score:} Benchmarks with a Z-score above $-1\sigma$ are considered to be in agreement with the reference; those below this threshold are not. 

\end{enumerate}

By incorporating the natural distribution of benchmark agreement scores, this method ensures that the assessment of agreement is both context-sensitive and adaptive to changes in the benchmark landscape.
Furthermore, as more benchmarks are added, the distribution is updated, making the test increasingly reflective of the current landscape of benchmarks measuring the desired trait.

\paragraph{Use More Models and Sample Them Randomly} 
\benchvalidname{} based on a small set of models tends to have high variance, as shown in Figure~\ref{fig:n_models_matters}, where the standard deviation of results can reach $0.25$ with fewer models (\S\ref{n_models_matter}). To reduce this variability and enhance reliability, we recommend using at least 10 models, preferably more. A larger and more diverse sample provides a more representative evaluation, minimizing bias and improving result stability. While increasing the number of models does raise computational costs, our recommendation remains practical, given that most model benchmarks already evaluate a larger number of models.
These models should represent the entire spectrum of available models, including diverse sizes, architectures, and training methods. Aiming for a random selection ensures equal representation and minimizes bias. Table~\ref{tab:ablations} shows that using this methodology to select models decreases \benchvalidname{} variance by more than $30$\%.

\paragraph{Report Multiple Granularities}
Benchmark agreement varies significantly with the range of model qualities considered, as demonstrated in Figure~\ref{fig:cover} (\S\ref{subsec:Resolution matters}). For instance, agreement can be high across a broad range of models but low among top-ranked models, which can mislead benchmark consumers who seek fine-grained distinctions. To address this, we recommend reporting agreement scores at multiple resolutions (e.g., 5/10/20 contiguous models, averaging across groups when more models were sampled). This practice provides a more nuanced and complete picture, allowing users to make informed decisions based on their specific needs. 
This approach provides a more nuanced view of benchmark agreement, highlighting critical distinctions that might otherwise be missed (e.g. the top 3 models are almost never in agreement across benchmarks).

\paragraph{Follow The Above Rules!}

Properly performing \benchvalidname{} using the above guidelines is not a trivial task. These methodologies require complex statistical tools, reproducible analysis and mostly, access to a large amount of up-to-date benchmarks data. 
Recognizing this difficulty, we have implemented our recommended workflow into \benchbench{}, a Python package for \benchvalidname{}, described below. 

Making the case for our above recommendations, Table~\ref{tab:ablations} demonstrates the significant gains obtained when using our methodological choices to perform \benchvalidname{}. 
It shows not only that the different recommendations each have an impact on variance, but also that their effect can be combined to achieve a substantially lower variance point -- reducing the standard deviation by $\sim67$\%, and thereby delivering far more robust \benchvalidname{} results.

\begin{figure*}[ht!]
    
    \includegraphics[width=\textwidth]{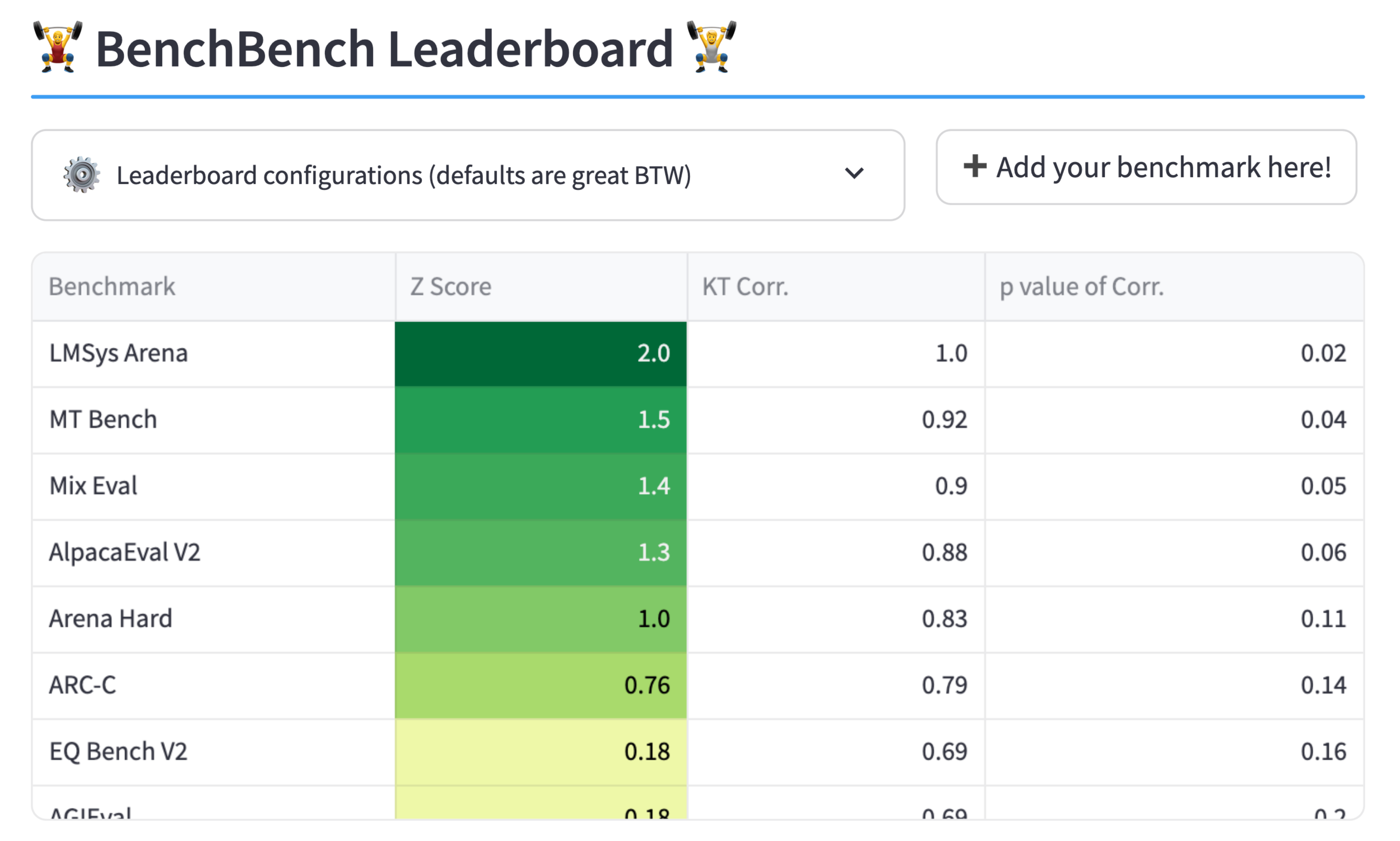}
    \caption{\textbf{The \benchbench{}-leaderboard - A meta-benchmark for \benchvalidname{}.} The following leaderboard is obtained with the default configurations, using the aggregate of all holistic reference benchmarks as the reference benchmarks and comparing subsets of 20 models  that were sampled randomly. As more benchmarks are added to Holistic set, results may be different upon view.}
    \label{fig:benchbench-leaderboard}
\end{figure*}

\section{\benchbench{} - a Package and Leaderboard}\label{sec:bb}
We introduce \benchbenchbold{}, a package implementing the above guidelines - standardizing the practice of \benchvalidname{} -- and holding results of multiple benchmarks for a wide variety of reference benchmark choices. The python package is available in GitHub at: \url{github.com/IBM/benchbench}.

The workflow of using the package is as follows:
\begin{enumerate}
\item A user enters their \benchvalidname{} configuration, including the desired group of reference benchmarks. 
\item \benchbench{} recommends a set of models for evaluation on the target benchmark.
\item The user inputs their benchmark results for the recommended models.
\item \benchbench{} produces a full \benchvalidname{} report.
\end{enumerate}

In the default functionality, \benchbench{} expects a list of model scores over the target benchmark, as well as a desired group of reference benchmarks to compare to. 
It also offers the functionality of proposing a minimal set of models for evaluation, ensuring fair and unbiased comparisons. 
While offering flexibility to change the defaults, \benchbench{}'s \benchvalidname{} report includes several granularities of models. \benchbench{} standardizes arbitrary decisions that hinder reproducibility, following the best practices proposed here.
Lastly, \benchbench{} offers the user to upload their benchmark results to the \benchbench{} database, enriching the reference benchmark distribution for future efforts, thereby enhancing \benchvalidname{} reliability without additional computational costs. due to running additional reference benchmarks.

We propose the \benchbenchbold{}\textbf{-leaderboard}, a new leaderboard designed to rank benchmarks according to their agreement to a desired group of reference benchmarks (see Figure~\ref{fig:benchbench-leaderboard}). 
To do so \benchbench{} ranks all submitted benchmarks by comparable standards. 

Since the \benchbench{}-leaderboard is build on top of the \benchbench{} package, new benchmarks uploaded to the package will be added to the leaderboard as well. Thus, the benchmark will improve with time, taking into account novel benchmarks and measured model traits.


\section{\benchvalidname{} uses in Related Work}

While some examples were given in the text, we elaborate on a handful of works employing \benchvalidname{}. 

Some works survey and analyze a field by utilizing \benchvalidname{} techniques.
\citet{Liu2021DoQA} check agreement across many QA datasets and conclude that since agreement is high, there is no need for more QA datasets. \citet{sun2023validity} use correlations to show that Compositionality Benchmarks do not agree amongst themselves. They used Kendall-Tau and set $0.7$ as the high agreement threshold.
Other works performed general efficient evaluation research and utilized \benchvalidname{} \citep{Prabhu2024LifelongBE,Perlitz2023EfficientB,Polo2024tinyBenchmarksEL,Viswanathan2023Prompt2ModelGD}. All of these works performed a thoughtful evaluation and large (reliable) rank correlation over all the models in the benchmarks. However, they did not consider the high correlations achieved in such settings (\S\ref{subsec:model_selection}).

Other work relies on \benchvalidname{} to compare to a specific benchmark.
\citet{Feng2024SampleEfficientHE} automatically sample a small set of instructions as an efficient LLM benchmark, reducing human labor significantly. They show this still agrees with existing benchmarks. Similarly, \citet{Lei2023S3EvalAS} and \citet{Viswanathan2023Prompt2ModelGD} both propose a synthetic benchmark as a proxy and show good agreement with the original benchmark, although they differ in their methodology.
\citet{Chang2023DoLM} propose two benchmarks and use agreement to show that they capture the same phenomenon, and \citet{Mizrahi2023StateOW} test agreement within the same benchmarks using different prompts.
\citet{Li2024TreeEvalBE} validate a new benchmark with 6 models of 3 sizes 7B,13B,33B with agreement alpaca(v2)~\citep{alpaca_eval}. 
\citet{Yuan2024PRobELMPR} and \citep{waldis2024holmes} show divergent validity by comparing their benchmark to established ones, showing low \benchvalidname{} scores.
Lastly, \citep{Perlitz2023EfficientB} compared efficient versions of the HELM benchmark to the full one.


\section{Discussion and Conclusions}

In this work, we shine a light on the lack of consistent \benchvalidname{} methodology. We analyze several \benchvalidname{} choices on a broad spectrum of benchmarks and assess their effect. Our analysis shows that different choices of (1) Models (2) Reference Benchmark(s), and (3) Thresholding scheme, can significantly alter \benchvalidname{} conclusions. Therefore, we advise a set of best practices and provide a Python package that aims to facilitate a consistent BAT process in the community. We also release the \benchbench{}-leaderboard, a benchmark that quantifies the agreement of a benchmark with an aggregate of existing benchmarks.

In this paper, our focus was on the methodological issues when performing \benchvalidname{}. We did not deal with questions regarding when \benchvalidname{} should be used, and how conclusions from \benchvalidname{} should be interpreted. Next, we describe several such open questions.

\paragraph{What do we make of high agreement?}
It is not trivial how one should treat two benchmarks that are in high agreement with each other. If one is more convenient to run (e.g., doesn't require costly metrics), then from a practical perspective, a user can simply choose it over the more expensive one. 
However, practitioners and researchers must not confuse high agreement with the notion that the benchmarks actually measure the exact same qualities. 
Among other things, this could lead to the erroneous conclusion that new benchmarks are no longer needed, impeding new benchmark development. 
The community must also discriminate between correlations of model abilities (strong models are strong at many tasks) and correlations of the benchmarks themselves (the benchmarks actually measure the same qualities).

\paragraph{What do we make of low agreement?}
Reliability concerns the consistency of benchmark results. 
In this paper, we accept the benchmark scores as presented and focus on their benchmark validity, which assesses whether benchmarks accurately measure what they purport to evaluate. 
However, this ignores the \textit{reliability} issues within the benchmarks, which place an upper bound on the level of benchmark agreement. 
If, for instance, a benchmark cannot reliably differentiate between its top-3 models, then naturally we do not expect to see agreement over the top-3 models with other benchmarks. 
Looking forward, methodological improvements in \benchvalidname{} must include incorporating reliability measures, allowing to decouple disagreements from low reliability. 

\paragraph{How do we use \benchvalidname{} to retire benchmarks?}
Another point concerns the role of \benchvalidname{} for benchmark retirement, i.e., at what point do we decide that an old benchmark is no longer relevant and should be discarded. Currently the issue of retirement is viewed mainly from the perspective of saturation, where the community stops using benchmarks on which all new models succeed. 
However, another reason to retire benchmarks may be that the mixture of abilities models are expected to possess has shifted over time. In this scenario, \benchvalidname{} can reveal that a certain benchmark is no longer viable.\\

In conclusion, our study enhances the precision and reliability of Benchmark Agreement Testing by establishing best practices and introducing the \benchbench{} Python package and leaderboard. These contributions foster standardized evaluations, enabling more accurate comparisons across benchmarks and setting a new direction for computational linguistics research.




\section{Limitations}

We note that finding low agreement may indicate one of two issues, both of which have negative implications. These issues should be addressed or interpreted differently. One option is that the benchmark measures something different from what it is supposed to and is hence not valid. That is the more common interpretation and calls for changes. Another option might be that the benchmark is just not reliable, intuitively its ranking is unstable and did not converge. In such cases, even the same benchmark may not agree with itself given small changes (subsets, seeds etc.), this usually calls for evaluating on more examples \citep{choshen-etal-2024-navigating} or configuration \citep{bandel2024unitxt}. There is a positive note to the same story, if a benchmark already shows a strong \benchvalidname{} in fine-grained evaluation (e.g., 5 models close to each other), it also means that it is quite reliable. 

Sometimes \benchvalidname{} is not needed. \benchvalidname{} gives a way to validate a benchmark by an external source of authority. However, other methods or other sources for authority (e.g., being masterfully crafted by experts) might give stronger signals. Especially in the case of new and unique signals that can mostly show they are different, but not that they are valid for their own unique purpose.

In general, \benchvalidname{} needs a reference benchmark, or ideally multiple benchmarks that provide diverse measurements of the same construct. Still, choosing the right reference benchmarks might be tricky, and the results might be sensitive to this choice.

%% file: ablation_table.tex
\begin{table}

\caption{\textbf{Our recommendations substantially reduce the variance of \benchvalidname{}.} Ablation analysis for each \benchvalidname{} recommendation separately and their combination. It shows great gains in using our methodologies when running \benchvalidname{} both separately and combined.}
\resizebox{\columnwidth}{!}{%
\begin{tabular}{ccc|cc|c}
\multicolumn{3}{c|}{Recommendations}                                                                                                                                             & \multicolumn{2}{c|}{BAT Variance}     & \multirow{2}{*}{\begin{tabular}[c]{@{}c@{}}Section\\ Ref.\end{tabular}}      \\
\begin{tabular}[c]{@{}c@{}}Aggregate\\ References\end{tabular} & \begin{tabular}[c]{@{}c@{}}Select\\ Metric\end{tabular} & \begin{tabular}[c]{@{}c@{}}Select\\ Models\end{tabular} & $\sigma$ ($\downarrow$)                  & Reduction &                                                                              \\ \hline
                                                              &                                                         &                                                        & \ccell{0.31} & -         & -                                                                            \\ \cline{1-3}
X                                     &                                                         &                                                        & \ccell{0.23} & $-30$\%   & \S\ref{subsec:reference_selection_matters} \\ \cline{1-3}
                                                              & X                               &                                                        & 0.23 & $-30$\%   & \S\ref{subsec:corr_metric_matters}         \\ \cline{1-3}
                                                              &                                                        & X                              & \ccell{0.20} & $-35$\%   & \S\ref{subsec:model_selection}              \\ \cline{1-3}
X                                     & X                               & X             & \ccell{0.10} & $-67$\%   & \S\ref{sec:best_practices}                 
\end{tabular}
}
\label{tab:ablations}
\end{table}


%% file: appendix.tex
\section{Appendices}

\subsection{Benchmarks used} \label{app:benchmark_used}

The \textbf{AGI Eval}~\citep{Zhong2023AGIEvalAH} benchmark assesses models on human-level cognition and problem-solving tasks, which tests the real-world applicability of model outputs. Similarly, \textbf{Alpaca (v2)}~\citep{alpaca_eval} and its \textbf{length-adjusted version}~\citep{Dubois2024LengthControlledAA} focus on a model's ability to follow complex instructions with the latter specifically addressing biases associated with output length.

\textbf{HumanEval}~\citep{Chen2021EvaluatingLL} presents code generation challenges, evaluating the syntactic correctness and logical soundness of model-generated code. Alongside, the \textbf{HuggingFace OpenLLM Leaderboard}~\citep{open-llm-leaderboard} employs the Eleuther AI Evaluation Harness~\citep{eval-harness} to test models on several key benchmarks such as ARC~\citep{clark2018think}, HellaSwag~\citep{zellers2019hellaswag}, MMLU~\citep{hendrycks2021measuring}, TruthfulQA~\citep{lin2022truthfulqa}, Winogrande~\citep{sakaguchi2021winogrande}, and GSM8k~\citep{cobbe2021gsm8k}. 
\textbf{EQ-Bench (v2)}~\citep{paech2023eqbench}, measures the emotional intelligence of models, essential for applications that involve nuanced human interactions.

The \textbf{MAGI}~\citep{MAGIbenchmark} benchmark integrates challenging elements from MMLU and AGIEval to test complex reasoning and problem-solving capabilities of models. It is particularly effective in highlighting subtle performance differences among top-tier models. 
\textbf{MMLU}~\citep{Hendrycks2020MeasuringMM} assesses both general and specialized knowledge across various domains, providing a broad evaluation spectrum.

Further, benchmarks like \textbf{Chatbot-Arena and MTBench}~\citep{Zheng2023JudgingLW} focus on multi-turn conversation abilities, crucial for applications in customer service and virtual assistance. 
Lastly, \textbf{Big Bench Hard}~\citep{suzgun2022challenging} challenges models with complex text understanding and generation, pushing the limits of what natural language processing technologies can achieve.
It is worth noting, that the HELM benchmark~\citep{liang2023holistic} was excluded from our analysis because there were few overlapping models with the other benchmarks.

\subsection{Model Tier}\label{subsec:Model Tier Matters}

\begin{figure}
    \centering
        \centering
        \includegraphics[width=0.7\linewidth]{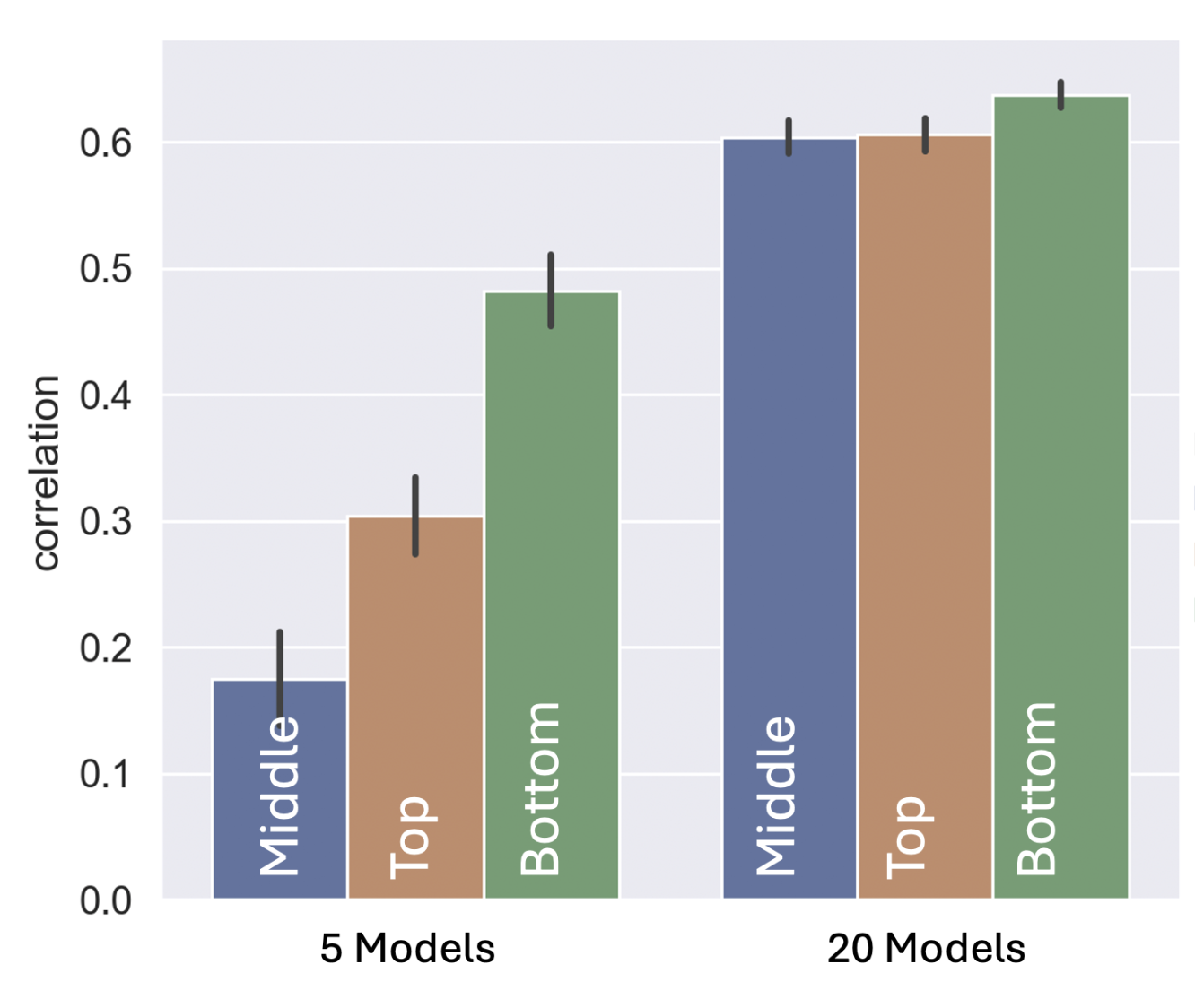} 
        \caption{\textbf{Correlation as a function of model subset size:} Correlations substantially decline as the models considered are closer to the top, error bars are the SEMs across the different pairs of benchmarks}
        \label{fig:top_middle_bottom_vs_n_models_used_bars}
\end{figure}

Building on the importance of model proximity, another crucial factor in benchmark agreement is the tier of models being assessed. Current \benchvalidname{} practices often treat benchmarks as a uniform slab, disregarding the variations across different tiers of model performance. However, agreement might not be uniform across these tiers, and understanding this variance can provide deeper insights into benchmark reliability and model performance.

In Figure~\ref{fig:top_middle_bottom_vs_n_models_used_bars}, we show that model tier significantly impacts benchmark agreement. Bottom-tier models exhibit higher agreement among themselves, with Kendall correlation coefficients just below 0.5. In contrast, middle-tier models show low agreement (coefficients below 0.2), and top-tier models demonstrate low to medium agreement (around 0.3).

One potential explanation for this phenomenon is the (lack of) reliability of the benchmark, as discussed in the introduction and literature \citep{Perlitz2023EfficientB}. Figure~\ref{fig:top_middle_bottom_vs_n_models_used_bars} highlights that the standard deviation of scores bottom-ranked models is significantly higher than the rest. This might mean that there is some effect the goes beyond granularity or density, with older models being easier to differentiate (and gaining higher correlations to the models). However middle and top ranked models do not show such a trend (even when taking into account that middle granularity is higher as top models are still joining the game), which means that no strong conclusion should be made excluding older models, switching benchmarks frequently or similar actions, at most, old models may be left out of \benchvalidname{}, but other effects seem more pressing.



\subsection{Benchmark used for visualizations}
\label{app:benchmarks_used_in_visualizations}

The benchmarks we used include: AGI Eval~\citep{Zhong2023AGIEvalAH}, Alpaca (v2)~\citep{alpaca_eval}, and its length-adjusted version~\citep{Dubois2024LengthControlledAA}, HuggingFace OpenLLM Leaderboard~\citep{open-llm-leaderboard}, MMLU~\citep{Hendrycks2020MeasuringMM}, Chatbot-Arena and MTBench~\citep{Zheng2023JudgingLW}, Big Bench Hard~\citep{suzgun2022challenging}.  ARC~\citep{clark2018think}, HellaSwag~\citep{zellers2019hellaswag}, TruthfulQA~\citep{lin2022truthfulqa}, Winogrande~\citep{sakaguchi2019winogrande}, 
EQ-Bench (v2)~\citep{paech2023eqbench}.
All benchmarks have a permissive license that allows academic use.